\documentclass[fleqn,10pt]{wlscirep}
\usepackage[utf8]{inputenc}
\usepackage[T1]{fontenc}
\usepackage{multirow}
\usepackage{nameref}
\usepackage{hyperref}
\usepackage{comment}
\usepackage{caption}
\usepackage{subcaption}
\usepackage{algorithm}
\usepackage{algpseudocode}
\title{Guilt Detection in Text: A Step Towards Understanding Complex Emotions}

\author[1,2]{Abdul Gafar Manuel Meque}
\author[1]{Nisar Hussain}
\author[1,*]{Grigori Sidorov}
\author[1]{Alexander Gelbukh}
\affil[1]{Instituto Politécnico Nacional (IPN), Centro de Investigación en Computación (CIC), Mexico City, Mexico }
\affil[2]{Catholic University of Mozambique, Faculdade de Economia e Gestao, Beira, 2100, Mozambique}

\affil[*]{sidorov@cic.ipn.mx}

\begin{abstract}
We introduce a novel Natural Language Processing (NLP) task called Guilt detection, which focuses on detecting guilt in text. We identify guilt as a complex and vital emotion that has not been previously studied in NLP, and we aim to provide a more fine-grained analysis of it. To address the lack of publicly available corpora for guilt detection, we created VIC, a dataset containing 4622 texts from three existing emotion detection datasets that we binarized into guilt and no-guilt classes. We experimented with traditional machine learning methods using bag-of-words and term frequency-inverse document frequency features, achieving a 72\% f1 score with the highest-performing model. Our study provides a first step towards understanding guilt in text and opens the door for future research in this area.
\end{abstract}

\begin{document}
\setcounter{secnumdepth}{2}
\flushbottom
\maketitle
\thispagestyle{empty}

\section*{Introduction}
In this study, we introduce Guilt detection, a novel task in Natural Language Processing aimed at detecting guilt in text. We also developed a dataset and a set of baseline classifiers for this task. Guilt is an emotion or feeling that arises when one contemplates past wrongdoings or thoughts regarding the morality of future actions, whether real or imagined. Throughout their lives, individuals experience a wide range of emotions and strongly need to connect with others and share their feelings with peers.

Sharing emotional experiences is a crucial component of the emotional process, as observed by \cite{rime1991}; furthermore, in \cite{bergermilkman2009}, the authors found that sharing  emotional experiences on the internet has become a crucial part of daily life on contemporary cultures. This can explain the popularity of emotion-related research in many scientific disciplines, including computer science, particularly Natural Language Processing (NLP), to determine the presence of, assess the intensity and polarity of, and compare various emotional experiences shared in the form of written text. Despite the considerable interest in emotion detection in general, there is still a sizable research vacuum regarding the in-depth examination of particular emotions, such as guilt.\\

 Guilt comes in many forms, namely anticipatory, existential, and reactive guilt \cite{rawlings1970a}, and it is frequently felt when people feel responsible for wrongdoing or  failing to uphold their own moral standards \cite{Kugler1992OnCA}, as a result, it frequently indicates an understanding of the experiences of others and a desire to correct any perceived interpersonal flaws \cite{treeby2021}. As with other self-conscious emotions, when in excess, guilt can have negative implications for one's mental health \cite{tangney2002a}, thus attracting the interest of researchers in various disciplines, from Psychology to neuroscience to Computer Science, in the latter, particularly as a sub-field in Natural Language Processing, guilt has been featured as a class in Emotion Detection sub-tasks in numerous NLP research works. \\
 
Considering the mental health implications of excessive guilt as mentioned in \cite{tangney2002a}, and numerous research studies have shown a link between guilt and suicidality in clinical populations, as discussed in \cite{treeby2021}, which cites two studies \cite{bryan2015} \cite{CUNNINGHAM2017227}, identifying guilt as one of the risk factors in suicidal ideation (SI), there is clear importance and value in studying ways to identify if, when and how an individual experiences guilt. Considering the finds of \cite{rime1991} and \cite{bergermilkman2009}, social media texts can be a great source for studying user-shared guilt experiences and feelings. However, to our knowledge, guilt has only been studied as a class in a multi-class emotion detection task and never as the primary subject of study in NLP, hence our interest in it. As a first step toward closing the research gap, we examined existing textual datasets from previous emotion detection tasks in which guilt was featured to build a binary guilt detection dataset. As detailed below, we intentionally selected data shared by the experiencers through different media and in different settings. In addition, we present a set of baseline models based on Traditional and Deep Learning methods. \\

The following are the main contributions of this paper:

\begin{itemize}
  \item A study of guilt detection from text using NLP techniques
  \item Development of a multi-source dataset for binary guilt detection, 
  \item Development of baseline models for the proposed dataset,
  \item An in-depth analysis of the dataset and models
\end{itemize}

The remainder of the paper is comprised of a \textbf{Literature Review} section showcasing works on emotion detection, especially those featuring guilt, \textbf{Dataset Development} section describing the techniques used in acquiring and building the dataset, \textbf{Benchmark Experiments} which present the baseline model experiment setups, followed by ”Results and discussion” where an initial analysis of the baseline results is made, concluding with \textbf{Conclusion and Future Work}, where we hint at possible future directions of our research work on guilt.

\section*{Literature Review}

To the best of our knowledge, there has not been any comprehensive research work done targeting Guilt, nor is there any dataset dedicated to guilt emotion, at least not one that is publicly available. Although  Guilt itself has not been the main subject of research, it has been featured in various studies, such as in \cite{balahur-etal-2011-detecting},  \cite{ghosh-etal-2020-cease}, where the authors explore a way and propose an approach to detect different types of emotions, using a Common Sense Knowledge EmotiNet, which they extended and improved in \cite{balahur-etal-2012-expanding}. \\
In this section, we present related work regarding both datasets and techniques for Emotion Detection in general and fine-grained detection of a specific emotion or emotion-related affective states featuring Guilt.\\

Due to its popularity, there exist a considerable number of datasets on Emotion Detection available publicly and featured in numerous studies, such as Vent \cite{Lykousas_2019}, XED \cite{ohman-etal-2020-xed}, GoEmotions \cite{demszky-etal-2020-goemotions}, EmotionLines  \cite{hsu-etal-2018-emotionlines}, Corpus of Emotion Annotated Suicide notes in English (CEASE) \cite{ghosh-etal-2020-cease}, The “International Survey on Emotion Antecedents and Reactions” (ISEAR) \cite{Scherer_1994}, deISEAR \cite{troiano2019crowdsourcing}, to mention a few. It is worth noting that many existing emotion detection datasets use emotion labels based on Ekman's \cite{ekman1971a, Ekman:1992a} or Plutchik's \cite{plutchik1980a} theory, or a variation thereof, often including the neutral label. In our study, we specifically focused on three datasets \cite{ghosh-etal-2020-cease, Lykousas_2019, Scherer_1994} that contain instances of guilt.

\textbf{Vent dataset}\cite{Lykousas_2019}, a large annotated dataset of texts (texts are  from 33M posts), emotions, and social connections from the Vent social media platform. In Vent, each post is associated with emotion, self-annotated with emotion by the post's author at the posting time. In \cite{Lykousas_2019} no experiments on ED were reported, but in \cite{Nurudin_2021} the authors ran ED experiments using Naive Bayes (NB), Random Forests (RF) and Logistic Regression (LR), Deep Neural Networks and Bi-LSTM. They employed two sets of feature representations, based on the type of method. For each of the traditional ML methods, they ran two sets of experiments, one using Bag-of-Words and then using TF-IDF feature vectors. The same procedure was used for Neural Network methods, but now using FastText and BERT embeddings as features. This resulted in ten (10) combinations of methods and features, with their best model achieving 19\% and 21\% in average macro and micro F1-score respectively, which is not surprising, considering the number of classes in Vent.

The \textbf{CEASE dataset} \cite{ghosh-etal-2020-cease}, consisting of 2393 sentences extracted from around 205 English language suicide notes, collected from various websites and annotated for 15 emotion classes. For their experiments, they first employed three deep learning (DNN) models (Convolutional Neural Network\cite{kim-2014-convolutional}, Long Short Term Memory\cite{hochreiter1997a}, and Gated Recurrent Unit model\cite{cho-etal-2014-properties}), the three DNN models were later combined to create two additional models (one majority vote ensemble and one multi-layer perceptron-based ensemble model). For comparison purposes, they trained and tested  various combinations of traditional ML classifiers (Multinomial-NB, RF, LR, and Support Vector Machine (SVC)) with different sets of features. The MLP ensemble and LTSM models achieved the best results, with an F1-Score of 59\% on average for all classes, and 48\% for \textbf{guilt}, 4 points less than the performance of the majority vote ensemble on this particular class. Among the traditional ML models, LR was the best-performing model, with an average F1-score of 48\% and 35\% for the guilt class.

The \textbf{ISEAR dataset} \cite{Scherer_1994}, a collection of 7669  sentences extracted from questionnaire answers from respondents from different countries, in different languages, as part of the ISEAR project in the '90s, the sentences were first translated and then annotated for 7 emotion classes. This dataset inspired the creation of two new datasets, the enISEAR (English ISEAR) deISEAR and (Deutsch ISEAR)\cite{troiano2019crowdsourcing}, both comprising 1001 descriptions obtained from the study participants after being presented with emotion and asked to describe an event in which they felt that particular emotion. Both datasets have the same number of descriptions, but are different in length, thus resulting in deISEAR being divided into 1084 sentences and a vocabulary size of 2613 distinct tokens, and enISEAR into 1366 sentences and a vocabulary of 3066. In \cite{troiano2019crowdsourcing} the authors trained a Maximmum Entropy (MaxEnt) classifier with unigram features on ISEAR\cite{Scherer_1994} and applied it to enISEAR and a translated version of deISEAR, achieving an average micro F1-Score of 47\%, on both datasets, 42\% on guilt class on deISEAR, and 41\% for the guilt class on enISEAR.

\section*{Dataset~ Development}

For this research, three existing datasets were used as a
starting point: the Vent \cite{Lykousas_2019},
ISEAR\cite{Scherer_1994} and CEASE \cite{ghosh-etal-2020-cease} ~Emotion datasets.
The selection of these specific datasets was mainly for two reasons, (1)
they are each from a different domain (source of where the texts were
collected), and (2) they all contain guilt as one of the classes.
A full description is provided below.

\subsection*{Dataset Preparation}

Initially, we considered all samples from each dataset because the Vent dataset contains 33M samples, we only selected samples from the feelings category, which is the category that contains the guilt subclass. This first sampling resulted in 4.358.680 samples from Vent, 7666 from ISEAR, and 2393 from CEASE. Table \ref{initial_sel_table} shows the emotion distribution based on this selection. Emotion classes that are present in at least two original data sources are labeled with the original name, and
those only appearing in one data source are labeled as ``others'' in
the table. As shown in Table~{\ref{initial_sel_table}}, there were a
total of ~4,368,739 samples, of which~135,601 were texts labelled with
guilt, and~4,233,138 with joy/happiness, anger, sadness/sorrow, fear,
disgust, shame, and other emotions.~

\begin{table}[!ht]
    \centering
    \begin{tabular}{|l|l|l|l|}
    \hline
        ~ & \multicolumn{3}{c|}{Dataset}   \\ \hline
        Emotion & Vent & ISEAR & CEASE \\ \hline
        guilt & 134434 & 1093 & 74 \\ \hline
        joy/happiness & - & 1094 & 38 \\ \hline
        anger & - & 1096 & 79 \\ \hline
        sadness/sorrow & - & 1096 & 305 \\ \hline
        fear & - & 1095 & 29 \\ \hline
        disgust & - & 1096 & - \\ \hline
        others & 4224246 & - & 1868 \\ \hline
        shame & - & 1096 & ~ \\ \hline
        Total & 4358680 & 7666 & 2393 \\ \hline
    \end{tabular}
    \caption{{\label{initial_sel_table}}Initial Data sampling result}
\end{table}

~With the initial selection, we proceeded with a second round of selection,
to deal with the enormous differences in size. For this, we first limited
the choice of non-guilt samples to the number of guilt samples found
in each original dataset, giving us 135,604 guilt and~135,604 not-guilt
samples, totaling 271.208 samples.

Because we are dealing with data from three different sources, we consider
a few different steps for cleaning the data, taking into account its
origin. The following steps were taken to identify and fix problems with the
dataset:

The vent dataset comes from automatically scraping millions of social
media posts, so it stands to reason that there might be some issues with
some of the samples, and a quick look into the texts, revealed some
samples consisting of only a single character space, special character,
or emojis, which are not helpful to our task. The second step was to
identify duplicated text-label combinations, followed by text
language identification. Since ISEAR was compiled from a questionnaire
answers, we check whether there are samples with no ``no response'',
``blank'' or similar values, which are a clear indication of the absence
of an answer from the respondents. As an extra cleaning step for the ISEAR,
we also checked for samples with ``No response.''

\par\null

\subsection*{VIC Dataset Statistics}
With the dataset fairly cleaned, we proceeded to sample all guilt
instances from each dataset and set the returned number of samples as a
cap for selecting non-guilt samples, which resulted in 245,826 samples,
which were then binarized (all non-guilt labels converted to ``no
guilt''), followed by a random selection of up to 1200 samples per class
per origin, which led to the final dataset comprised of 4622 samples,
with a balanced class distribution.~

\begin{figure}
    \centering

    \includegraphics[width=\textwidth]{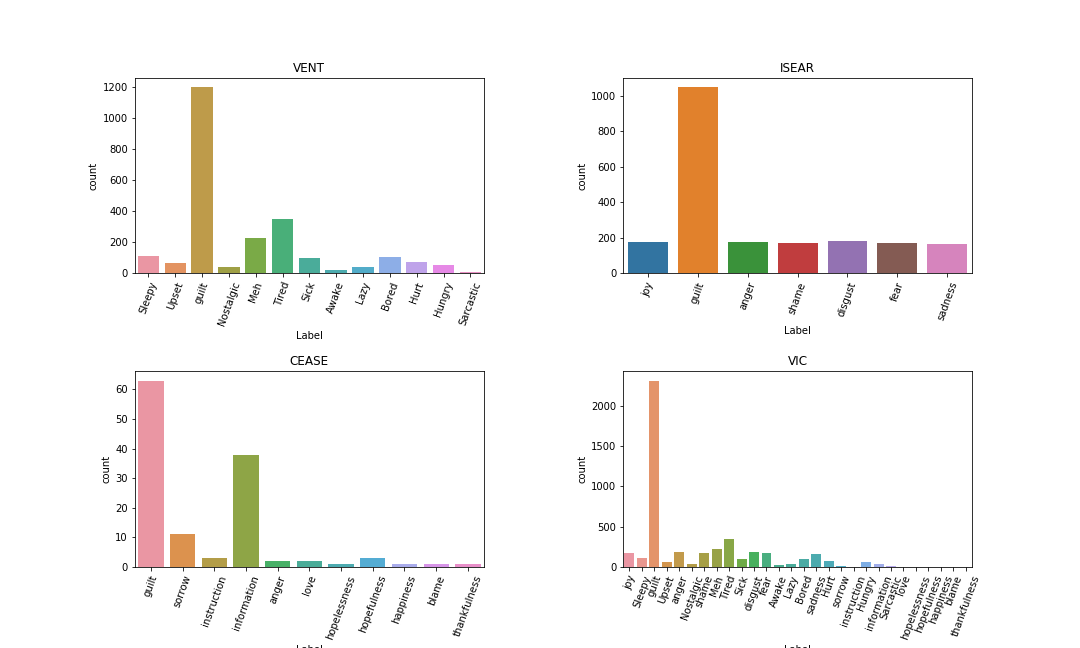}
    \caption{Original label distribution for each subset}
    \label{fig:original_label_dist}
\end{figure}

% \usepackage[normalem]{ulem}
\begin{table}[!ht]
\centering
\resizebox{0.87\columnwidth}{!}{%
 \begin{tabular}{p{0.60\linewidth} | p{0.15\linewidth}| p{0.20\linewidth}}
\textbf{Text} & \textbf{Origin Label} & \textbf{Binary Label} \\  \hline
why do i always go back to christian  why does he accept every time i fuck up and go back to him & guilt & guilt \\  \hline
its two am and all i can think about is you i want you here i need you now i dont think i can ever walk away from you  i wish i could move on but how do you move on from something youve never really had ugh & Tired & no guilt \\  \hline
my mom brought me and sister home because we were hungry and tired so i'm back in my nice bed & Sleepy & no guilt \\  \hline
vent would be more useful if i could actually use words to describe my feelings & Tired & no guilt \\  \hline
on embarking on university life i came from a different city and did not know anybody at the uni i was frightened because my well known and loved friends also all my security had been taken away & fear & no guilt \\  \hline
fear  paralysing  that i would not be accepted by the god who i believed to be there because i was morally bankrupt before becoming a christian and realising that that was why christ came to free us from sin and to forgive us & fear & no guilt \\  \hline
i am sorry to the people that i love but i can not fucking take it anymore & guilt & guilt \\  \hline
on a day like this everyone seeks the company of beloved ones & love & no guilt \\  \hline
i am sorry once again & guilt & guilt \\  \hline
i angered a close friend and he was injured & guilt & guilt \\  
\end{tabular}
}
\caption{ A sample of texts and labels, with origin label from dataset VIC}
\label{tab:vic_sample_texts}
\end{table}
Our resulting dataset (VIC) contains 126 samples from CEASE, 2,096 from ISEAR, and 1248 from Vent, Table \ref{tab:vic_sample_texts} presents a sample of texts in VIC and Table \ref{tab:vic_stats} details the label distribution by sample origin and Figure \ref{fig:original_label_dist} depicts the distribution of non-guilt class samples distribution for each origin and their distribution in VIC. As shown in Table \ref{tab:vic_stats}, initial data exploration revealed that guilt samples generally have a slightly smaller average word length when compared to the non-guilt samples, and the largest texts are of non-guilt samples.

\begin{table}[!ht]
\centering
\begin{tabular}{|l|l|l|}
\hline
\textbf{Class}    & \textbf{\# of Samples} & \textbf{Sample Origin} \\ \hline
\textbf{Guilt}    & 63                     & \multirow{2}{*}{CEASE} \\ \cline{1-2}
\textbf{No Guilt} & 63                     &                        \\ \hline
\textbf{Guilt}    & 1048                   & \multirow{2}{*}{ISEAR} \\ \cline{1-2}
\textbf{No Guilt} & 1048                   &                        \\ \hline
\textbf{Guilt}    & 1200                   & \multirow{2}{*}{Vent}  \\ \cline{1-2}
\textbf{No Guilt} & 1048                   &                        \\ \cline{1-2} \hline
\end{tabular}
\caption{A summary of label and origin distribution in the resulting dataset VIC}
\label{tab:vic_stats}
\end{table}

\section*{}

{\label{185795}}

\section*{Benchmark Experiments} 
{\label{3344}}

To provide a baseline for the guilt detection task, we conducted two sets of experiments, one using traditional machine learning and the second using  neural network-based methods. For the traditional machine learning methods, we first conducted hyperparameter tuning using a Grid Search algorithm and then trained and tested models with the best parameters from the grid search which was then compared with the results from neural networks models without tuning them. The full details of the experiments are detailed in the following subsection. To properly assess the effects of the dataset origin in the task, for each combination of method and features, we tested on subsets of the dataset based on the sample origin and finally on the whole dataset.

\subsection*{Methods} \label{sec:methods}

\textbf{Traditional Machine Learning:}~we have chosen three methods, namely Support Vector Machine (SVM) a machine learning algorithm widely used in classification tasks. It has the property of being able to learn the correct labeling of data points that belong to different classes, given a set of training examples. Multinomial Naive Bayes, a Bayesian learning approach also widely used in natural language processing tasks, and Logistic Regression, a binary classifier that uses sigmoid activation in the output layer to predict the label. For each method, we experimented using two different features type, BoW and tf-idf vectors.
\\
\textbf{Neural Networks:~} for neural network methods, we selected Convolutional Neural Networks, and inspired by the experiments in \cite{fazl_etal_hope_2022} and \cite{fazl_etall_guilt_2022}, we added Bi-directional Long Short Term Memory (BiLSTM) to our choices of neural network models. All of our Sequential models start with an Embedding layer, with 64 dimensions and the vocabulary and input length based on the training data, with a final Dense layer with sigmoid activation.
\begin{itemize}
    \item CNN is a feed-forward neural network called commonly used to analyze visual images by processing data in a grid-like architecture. According to \cite{fazl_etall_guilt_2022} CNNs are able to automatically learn features from the input text and have been successful in many natural language processing tasks. in addition to the layers in common to all the other models, our CNN has a 1-D Convolutional layer, a Global Max Pooling layer, and a Dense layer with relu activation.

    \item  Bi-LSTM \cite{Schuster_1997}, a form of recurrent neural network that uses LSTM cells (i.e. cell state representations) to adaptively change the size of its step length between input. This mechanism makes it possible for the network to learn long and short-term dependencies, as well as dynamics that span multiple time steps and hidden states. this NN is made of a Bidirectional LSTM layer and Dropout layers.

\end{itemize}

\subsection*{Feature engineering}

\textbf{Bag of Words}: each instance of text in our corpus is treated as
a collection of words, that is then vectorized using a Count Vectorizer
using sklearn library's feature extraction package, with  n=\{1-4\}.\\
\textbf{TFIDF}: as a second feature option, we used the n-gram term
frequency and inverse document frequency Vectors with n=\{1-4\},

\subsection*{Hyperparameter tuning with GridSearch~}

We used TF-IDF as features and trained classical ML classifiers for the first set of experiments. We used GridSearch to find the best hyperparameters (C for SVM and LR and alpha for MNB for example) and feature combinations for each ML model. The entire hyperparameter search space is presented in Table \ref{tab: hyperparameter-search-space}.

\begin{table}[!ht]
\centering
\resizebox{0.52\columnwidth}{!}{%
\begin{tabular}{|c|l|l|}
\hline
\multicolumn{1}{|l|}{\textbf{Method}} & \textbf{Hyperparameter} & \textbf{Search Space}                                    \\ \hline
\multirow{3}{*}{\textbf{LR}}          & solver                  & {[}newton-cg,   lbfgs,sag, liblinear, saga{]}            \\ \cline{2-3} 
                                      & C                       & {[}0.01, 0.1, 1, 10, 100{]}                              \\ \cline{2-3} 
                                      & penalty                 & {[}'l1', 'l2', 'elasticnet'{]}                           \\ \hline
\multirow{3}{*}{\textbf{SVC}}         & kernel                  & linear, rbf{]}                                           \\ \cline{2-3} 
                                      & C                       & {[}0.01, 0.1, 1, 10, 100{]}                              \\ \cline{2-3} 
                                      & gamma                   & {[}0.001, 0.0001{]}                                      \\ \hline
\multirow{2}{*}{\textbf{GB}}          & n\_estimators           & {[}16, 32{]}                                             \\ \cline{2-3} 
                                      & learning\_rate          & {[}0.8, 1.0{]}                                           \\ \hline
\multirow{3}{*}{\textbf{DT \& RF}}    & max\_depth              & {[}5,10,25,None{]}                                       \\ \cline{2-3} 
                                      & min\_samples\_leaf      & {[}5, 10, 20, 50, 100{]}                                 \\ \cline{2-3} 
                                      & criterion               & {[}gini, entropy{]}                                      \\ \hline
\textbf{RF}                           & n\_estimators           & {[}16, 32{]}                                             \\ \hline
\multirow{3}{*}{\textbf{MNB}}         & alpha                   & {[}0.1, 0.2, 0.3, 0.4, 0.5, 0.6, 0.7, 0.8,   0.9, 1.0{]} \\ \cline{2-3} 
                                      & fit\_prior              & {[}True, False{]}                                        \\ \cline{2-3} 
                                      & class\_prior            & {[}None, {[}.1,.9{]},{[}.2, .8{]}{]}                     \\ \hline
\end{tabular}
 }
\caption{Hyper parameter search configuration}
\label{tab: hyperparameter-search-space}
\end{table}

\subsection*{Dataset experiment setup} \label{subsec:datasetsplit}

\textbf{Origin-based subsetting :} We selected the samples from VIC based
on their origin, resulting in three sub-datasets, and ran experiments on each of them separately, using the standard train and test split, with 10\% samples for testing and 90\% for training each time.\\
\textbf{Combined and Shuffled:~}with the help of the pandas library, we shuffled the entire VIC dataset and conducted experiments on it, also using the 90-10 train and test splits.\\
\textbf{Leave-one-out Training and Testing:} In addition to the experiments detailed above, we ran train and test experiments using examples from two origins for training and the third one for testing.

\section*{Results and discussion}

We started by running a hyperparameter search for the traditional ML models (which resulted in 478 candidate model configurations, from combining methods, hyperparameters, and features) on the VIC dataset. The resulting models were then ranked based on their overall performance, and for each method, we selected their best parameters for the subsequent tests. As shown in Table \ref{tab:hyper-paramter-search-result}, for the Tfidf feature type, the best performing algorithm is MNB with an F1-score of 0.72, using alpha=0.9, class\_prior=None, and fit\_prior=True. LR and SVC also have similar performances, with F1 scores of 0.68 and 0.67, respectively. The best-performing RF and DT have F1 scores of 0.67 and 0.65, respectively.

For the Bow feature type, MNB also has the highest F1-score of 0.72, using alpha=0.7, class\_prior=None, and fit\_prior=True. LR has an F1-score of 0.69, while SVC has an F1-score of 0.67. The best-performing RF and DT have F1 scores of 0.67 and 0.66, respectively. The best-performing GB has an F1-score of 0.63, using n\_estimators=32 and lr=1.

Table \ref{tab:hyper-paramter-search-result}  shows the hyperparameter results for different machine learning algorithms and feature types. The evaluation metric used is F1-score, and the hyperparameters selected by the tuning process are listed as well.
 
\begin{table}[!ht]
\centering
\resizebox{0.84\columnwidth}{!}{
\begin{tabular}{|c|l|r|r|r|r|l|} \hline 
\textbf{Features} & \multicolumn{1}{c|}{\textbf{Estimator}} & \textbf{min} & \textbf{mean} & \textbf{max} & \textbf{std dev} & \textbf{Hyperparameters} \\ \hline 
\multirow{6}{*}{Tfidf} & Multinomial Naive Bayes & 0.71 & 0.72 & 0.75 & 0.015 & $\alpha=1.0$, class\_prior=None, fit\_prior=True \\ 
\cline{2-7} & Support Vector Machine & 0.68 & 0.69 & 0.71 & 0.009 & $C=100$, kernel=linear \\ 
\cline{2-7} & Logistic Regression & 0.68 & 0.69 & 0.70 & 0.007 & $C=100$, penalty=l2, solver=saga \\ \cline{2-7} 
& Random Forest & 0.33 & 0.58 & 0.67 & 0.127 & $n\_estimators=32$, criterion=gini, max\_depth=None, min\_samples\_leaf=100 \\ 
\cline{2-7} & Gradient Boosting & 0.60 & 0.62 & 0.64 & 0.014 & $n\_estimators=32$, lr=0.8 \\ 
\cline{2-7} & Decision Tree & 0.60 & 0.61 & 0.63 & 0.010 & criterion=entropy, max\_depth=None, min\_samples=10 \\ \hline 
\multirow{6}{*}{BoW} & Multinomial Naive Bayes & 0.70 & 0.71 & 0.73 & 0.011 & $\alpha=0.7$, class\_prior=\{0.1,0.9\}, fit\_prior=True \\ 
\cline{2-7} & Logistic Regression & 0.65 & 0.67 & 0.69 & 0.016 & $C=10$, penalty=l2, solver=sag \\ \cline{2-7} 
& Random Forest & 0.47 & 0.61 & 0.69 & 0.079 & $n\_estimators=32$, criterion=gini, max\_depth=None, min\_samples\_leaf=10 \\ 
\cline{2-7} & Support Vector Machine & 0.65 & 0.66 & 0.67 & 0.008 & $C=10$, kernel=linear \\ 
\cline{2-7} & Gradient Boosting & 0.63 & 0.65 & 0.67 & 0.014 & $n\_estimators=32$, lr=0.8 \\ \cline{2-7} 
& Decision Tree & 0.52 & 0.57 & 0.62 & 0.034 & criterion=entropy, max\_depth=None, min\_samples=10 \\ \hline 
\end{tabular} 
}
\caption{Hyperparameter search results for different machine learning algorithms and feature types using F1-score as the evaluation metric.}
\label{tab:hyper-paramter-search-result}
\end{table}

\subsection*{Experiments on VIC dataset}

Based on the Hyperparameter search results detailed in Table \ref{tab:hyper-paramter-search-result}, we again trained and tested twelve (12) classical ML classifiers with their best-performing hyperparameters, using a K-Fold cross-validation strategy, with \textit{k=5}, thus training on 80\% of the data and testing on the remainder 20\%, on each fold. Multinomial Naive Bayes achieved the best performance, 72\% F-Score when trained using Tfidf vectors as features and 71\% with BoW feature vectors, as expected, as per the result of our hyperparameter tuning. Additionally, we ran tests using neural network classifiers, a CNN and BiLSTM as described in the previous section, equally trained and tested with the same K-Fold strategy as our classical ML models. Our CNN model achieved an F1-Score of 68\%, 4\% less than our best-performing MNB, which can be attributed at the fact that the proposed NN models were not subject to any fine-tuning.  Regardless of the feature used, RF models had the worse performance amongst all the classifiers we experimented on, with an F1-Score of 52\% overall. In general, Tfidf feature vectors gave better results for all methods when compared with BoW which is in line with the findings of previous research, with the only exception being the RF classifier, and the detailed results are depicted in Table \ref{tab:VICExperiments}.

\begin{table}[!ht]

\centering
\resizebox{0.42\columnwidth}{!}{%
\begin{tabular}{|llllll|}
\hline

\multicolumn{1}{l|}{\textbf{Feature}} & \multicolumn{1}{l|}{\textbf{Model}} & \multicolumn{1}{l|}{\textbf{Acc}} & \multicolumn{1}{l|}{\textbf{Precision}} & \multicolumn{1}{l|}{\textbf{Recall}} &\multicolumn{1}{l}{ \textbf{F1-Score}} \\ \hline
\multicolumn{6}{c}{\textbf{Traditional Machine Learning Models}} \\ \hline
\multicolumn{1}{|c|}{\multirow{6}{*}{\textbf{tfidf}}} & \multicolumn{1}{l|}{\textbf{DT}} & \multicolumn{1}{l|}{0.61} & \multicolumn{1}{l|}{0.62} & \multicolumn{1}{l|}{0.61} & 0.61 \\ \cline{2-6} 
\multicolumn{1}{|c|}{} & \multicolumn{1}{l|}{\textbf{RF}} & \multicolumn{1}{l|}{0.52} & \multicolumn{1}{l|}{0.53} & \multicolumn{1}{l|}{0.75} & 0.58 \\ \cline{2-6} 
\multicolumn{1}{|c|}{} & \multicolumn{1}{l|}{\textbf{MNB}} & \multicolumn{1}{l|}{0.68} & \multicolumn{1}{l|}{0.64} & \multicolumn{1}{l|}{\textbf{0.83}} & \textbf{0.72} \\ \cline{2-6} 
\multicolumn{1}{|c|}{} & \multicolumn{1}{l|}{\textbf{LR}} & \multicolumn{1}{l|}{\textbf{0.70}} & \multicolumn{1}{l|}{\textbf{0.72}} & \multicolumn{1}{l|}{0.67} & 0.69 \\ \cline{2-6} 
\multicolumn{1}{|c|}{} & \multicolumn{1}{l|}{\textbf{GB}} & \multicolumn{1}{l|}{0.65} & \multicolumn{1}{l|}{0.68} & \multicolumn{1}{l|}{0.56} & 0.61 \\ \cline{2-6} 
\multicolumn{1}{|c|}{} & \multicolumn{1}{l|}{\textbf{SVC}} & \multicolumn{1}{l|}{\textbf{0.70}} & \multicolumn{1}{l|}{\textbf{0.72}} & \multicolumn{1}{l|}{0.67} & 0.69 \\ \hline
\multicolumn{1}{|c|}{\multirow{6}{*}{\textbf{BoW}}} & \multicolumn{1}{l|}{\textbf{DT}} & \multicolumn{1}{l|}{0.62} & \multicolumn{1}{l|}{0.63} & \multicolumn{1}{l|}{0.59} & 0.61 \\ \cline{2-6} 
\multicolumn{1}{|c|}{} & \multicolumn{1}{l|}{\textbf{RF}} & \multicolumn{1}{l|}{0.53} & \multicolumn{1}{l|}{0.55} & \multicolumn{1}{l|}{0.59} & 0.52 \\ \cline{2-6} 
\multicolumn{1}{|c|}{} & \multicolumn{1}{l|}{\textbf{MNB}} & \multicolumn{1}{l|}{0.66} & \multicolumn{1}{l|}{0.62} & \multicolumn{1}{l|}{\textbf{0.83}} & \textbf{0.71} \\ \cline{2-6} 
\multicolumn{1}{|c|}{} & \multicolumn{1}{l|}{\textbf{LR}} & \multicolumn{1}{l|}{0.66} & \multicolumn{1}{l|}{0.66} & \multicolumn{1}{l|}{0.69} & 0.67 \\ \cline{2-6} 
\multicolumn{1}{|c|}{} & \multicolumn{1}{l|}{\textbf{GB}} & \multicolumn{1}{l|}{0.65} & \multicolumn{1}{l|}{\textbf{0.70}} & \multicolumn{1}{l|}{0.52} & 0.59 \\ \cline{2-6} 
\multicolumn{1}{|c|}{} & \multicolumn{1}{l|}{\textbf{SVC}} & \multicolumn{1}{l|}{\textbf{0.68}} & \multicolumn{1}{l|}{\textbf{0.70}} & \multicolumn{1}{l|}{0.63} & 0.67 \\ \hline
\multicolumn{6}{c}{\textbf{Neural Network Models}} \\ \hline

\multicolumn{1}{|c|}{\multirow{2}{*}{\textbf{text tokens}}} & \multicolumn{1}{l|}{CNN} & \multicolumn{1}{l|}{\textbf{0.68}} & \multicolumn{1}{l|}{\textbf{0.67}} & \multicolumn{1}{l|}{\textbf{0.70}} & \textbf{0.68} \\ \cline{2-6} 
\multicolumn{1}{|c|}{} & \multicolumn{1}{l|}{bilstm} & \multicolumn{1}{l|}{0.64} & \multicolumn{1}{l|}{0.64} & \multicolumn{1}{l|}{0.66} & 0.64 \\ \hline
\end{tabular}
}
\caption{Detailled results of different combinations of ML methods and features on the entire VIC dataset}
\label{tab:VICExperiments}
\end{table}

\subsection*{Origin-based subsetting}

To better understand the effects of sample origin on the results, we conducted training and testing on each subset of our data based on the sample origin. For each method+feature combination, on each subset, we calculated the accuracy, precision, recall, and F1 scores, which are presented as summarized in Table~{\ref{tab:result_per_origin}}. This result tells us that:

\begin{table}[!ht]
\centering
\resizebox{0.82\columnwidth}{!}{%

\begin{tabular}{|clllllllllllll|}

\multicolumn{2}{c}{} & \multicolumn{4}{c}{\textbf{Vent}} & \multicolumn{4}{c}{\textbf{ISEAR}} & \multicolumn{4}{c}{\textbf{CEASE}} \\ \hline
\multicolumn{1}{|l|}{\textbf{Feature}} & \multicolumn{1}{c|}{\textbf{Model}} & \multicolumn{1}{l|}{\textbf{Acc.}} & \multicolumn{1}{l|}{\textbf{Prec.}} & \multicolumn{1}{l|}{\textbf{Rec.}} & \multicolumn{1}{l|}{\textbf{F1-Score}} & \multicolumn{1}{l|}{\textbf{Acc.}} & \multicolumn{1}{l|}{\textbf{Prec.}} & \multicolumn{1}{l|}{\textbf{Rec.}} & \multicolumn{1}{l|}{\textbf{F1-Score}} & \multicolumn{1}{l|}{\textbf{Acc.}} & \multicolumn{1}{l|}{\textbf{Prec.}} & \multicolumn{1}{l|}{\textbf{Rec.}} & \textbf{F1-Score} \\ \hline
\multicolumn{14}{c}{\textbf{Traditional Machine Learning Models}} \\ \hline
\multicolumn{1}{|c|}{\multirow{6}{*}{\textbf{Tfidf}}} & \multicolumn{1}{l|}{\textbf{DT}} & \multicolumn{1}{l|}{0.59} & \multicolumn{1}{l|}{0.59} & \multicolumn{1}{l|}{0.58} & \multicolumn{1}{l|}{0.59} & \multicolumn{1}{l|}{0.62} & \multicolumn{1}{l|}{0.63} & \multicolumn{1}{l|}{0.61} & \multicolumn{1}{l|}{0.62} & \multicolumn{1}{l|}{0.68} & \multicolumn{1}{l|}{0.73} & \multicolumn{1}{l|}{0.62} & 0.66 \\ \cline{2-14} 
\multicolumn{1}{|c|}{} & \multicolumn{1}{l|}{\textbf{RF}} & \multicolumn{1}{l|}{0.52} & \multicolumn{1}{l|}{0.54} & \multicolumn{1}{l|}{0.46} & \multicolumn{1}{l|}{0.45} & \multicolumn{1}{l|}{0.52} & \multicolumn{1}{l|}{0.22} & \multicolumn{1}{l|}{0.26} & \multicolumn{1}{l|}{0.24} & \multicolumn{1}{l|}{0.49} & \multicolumn{1}{l|}{0.40} & \multicolumn{1}{l|}{0.80} & 0.53 \\ \cline{2-14} 
\multicolumn{1}{|c|}{} & \multicolumn{1}{l|}{\textbf{MNB}} & \multicolumn{1}{l|}{0.66} & \multicolumn{1}{l|}{0.62} & \multicolumn{1}{l|}{0.82} & \multicolumn{1}{l|}{\textbf{0.71}} & \multicolumn{1}{l|}{0.72} & \multicolumn{1}{l|}{0.69} & \multicolumn{1}{l|}{0.82} & \multicolumn{1}{l|}{\textbf{0.75}} & \multicolumn{1}{l|}{0.77} & \multicolumn{1}{l|}{0.72} & \multicolumn{1}{l|}{0.91} & \textbf{0.80} \\ \cline{2-14} 
\multicolumn{1}{|c|}{} & \multicolumn{1}{l|}{\textbf{LR}} & \multicolumn{1}{l|}{0.68} & \multicolumn{1}{l|}{0.69} & \multicolumn{1}{l|}{0.65} & \multicolumn{1}{l|}{0.67} & \multicolumn{1}{l|}{0.73} & \multicolumn{1}{l|}{0.75} & \multicolumn{1}{l|}{0.71} & \multicolumn{1}{l|}{0.73} & \multicolumn{1}{l|}{0.75} & \multicolumn{1}{l|}{0.71} & \multicolumn{1}{l|}{0.84} & 0.77 \\ \cline{2-14} 
\multicolumn{1}{|c|}{} & \multicolumn{1}{l|}{\textbf{GB}} & \multicolumn{1}{l|}{0.64} & \multicolumn{1}{l|}{0.67} & \multicolumn{1}{l|}{0.55} & \multicolumn{1}{l|}{0.60} & \multicolumn{1}{l|}{0.67} & \multicolumn{1}{l|}{0.70} & \multicolumn{1}{l|}{0.61} & \multicolumn{1}{l|}{0.65} & \multicolumn{1}{l|}{0.79} & \multicolumn{1}{l|}{0.91} & \multicolumn{1}{l|}{0.65} & 0.75 \\ \cline{2-14} 
\multicolumn{1}{|c|}{} & \multicolumn{1}{l|}{\textbf{SVC}} & \multicolumn{1}{l|}{0.68} & \multicolumn{1}{l|}{0.69} & \multicolumn{1}{l|}{0.64} & \multicolumn{1}{l|}{0.67} & \multicolumn{1}{l|}{0.73} & \multicolumn{1}{l|}{0.74} & \multicolumn{1}{l|}{0.71} & \multicolumn{1}{l|}{0.73} & \multicolumn{1}{l|}{0.76} & \multicolumn{1}{l|}{0.84} & \multicolumn{1}{l|}{0.65} & 0.73 \\ \hline
\hline
\multicolumn{1}{|c|}{\multirow{6}{*}{\textbf{BoW}}} & \multicolumn{1}{l|}{\textbf{DT}} & \multicolumn{1}{l|}{0.62} & \multicolumn{1}{l|}{0.63} & \multicolumn{1}{l|}{0.60} & \multicolumn{1}{l|}{0.61} & \multicolumn{1}{l|}{0.65} & \multicolumn{1}{l|}{0.66} & \multicolumn{1}{l|}{0.63} & \multicolumn{1}{l|}{0.64} & \multicolumn{1}{l|}{0.72} & \multicolumn{1}{l|}{0.75} & \multicolumn{1}{l|}{0.69} & 0.71 \\ \cline{2-14} 
\multicolumn{1}{|c|}{} & \multicolumn{1}{l|}{\textbf{RF}} & \multicolumn{1}{l|}{0.52} & \multicolumn{1}{l|}{0.54} & \multicolumn{1}{l|}{0.42} & \multicolumn{1}{l|}{0.43} & \multicolumn{1}{l|}{0.51} & \multicolumn{1}{l|}{0.20} & \multicolumn{1}{l|}{0.15} & \multicolumn{1}{l|}{0.13} & \multicolumn{1}{l|}{0.49} & \multicolumn{1}{l|}{0.40} & \multicolumn{1}{l|}{0.80} & 0.53 \\ \cline{2-14} 
\multicolumn{1}{|c|}{} & \multicolumn{1}{l|}{\textbf{MNB}} & \multicolumn{1}{l|}{0.64} & \multicolumn{1}{l|}{0.61} & \multicolumn{1}{l|}{0.82} & \multicolumn{1}{l|}{\textbf{0.70}} & \multicolumn{1}{l|}{0.70} & \multicolumn{1}{l|}{0.66} & \multicolumn{1}{l|}{0.83} & \multicolumn{1}{l|}{\textbf{0.74}} & \multicolumn{1}{l|}{0.60} & \multicolumn{1}{l|}{0.56} & \multicolumn{1}{l|}{0.95} & 0.71 \\ \cline{2-14} 
\multicolumn{1}{|c|}{} & \multicolumn{1}{l|}{\textbf{LR}} & \multicolumn{1}{l|}{0.65} & \multicolumn{1}{l|}{0.64} & \multicolumn{1}{l|}{0.71} & \multicolumn{1}{l|}{0.67} & \multicolumn{1}{l|}{0.73} & \multicolumn{1}{l|}{0.73} & \multicolumn{1}{l|}{0.72} & \multicolumn{1}{l|}{0.73} & \multicolumn{1}{l|}{0.75} & \multicolumn{1}{l|}{0.74} & \multicolumn{1}{l|}{0.75} & \textbf{0.74} \\ \cline{2-14} 
\multicolumn{1}{|c|}{} & \multicolumn{1}{l|}{\textbf{GB}} & \multicolumn{1}{l|}{0.66} & \multicolumn{1}{l|}{0.72} & \multicolumn{1}{l|}{0.52} & \multicolumn{1}{l|}{0.61} & \multicolumn{1}{l|}{0.69} & \multicolumn{1}{l|}{0.73} & \multicolumn{1}{l|}{0.61} & \multicolumn{1}{l|}{0.66} & \multicolumn{1}{l|}{0.76} & \multicolumn{1}{l|}{0.90} & \multicolumn{1}{l|}{0.60} & 0.71 \\ \cline{2-14} 
\multicolumn{1}{|c|}{} & \multicolumn{1}{l|}{\textbf{SVC}} & \multicolumn{1}{l|}{0.66} & \multicolumn{1}{l|}{0.68} & \multicolumn{1}{l|}{0.60} & \multicolumn{1}{l|}{0.64} & \multicolumn{1}{l|}{0.73} & \multicolumn{1}{l|}{0.74} & \multicolumn{1}{l|}{0.71} & \multicolumn{1}{l|}{0.72} & \multicolumn{1}{l|}{0.73} & \multicolumn{1}{l|}{0.82} & \multicolumn{1}{l|}{0.59} & 0.68 \\ \hline
\multicolumn{14}{c}{\textbf{Neural Network Models}} \\ \hline

\multicolumn{1}{|c|}{\multirow{2}{*}{\textbf{text tokens}}} & \multicolumn{1}{l|}{\textbf{CNN}} & \multicolumn{1}{l|}{0.63} & \multicolumn{1}{l|}{0.68} & \multicolumn{1}{l|}{0.61} & \multicolumn{1}{l|}{\textbf{0.66}} & \multicolumn{1}{l|}{0.71} & \multicolumn{1}{l|}{0.71} & \multicolumn{1}{l|}{0.73} & \multicolumn{1}{l|}{\textbf{0.71}} & \multicolumn{1}{l|}{0.74} & \multicolumn{1}{l|}{0.74} & \multicolumn{1}{l|}{0.77} & \textbf{0.74} \\ \cline{2-14} 
\multicolumn{1}{|c|}{} & \multicolumn{1}{l|}{\textbf{BiLSTM}} & \multicolumn{1}{l|}{0.64} & \multicolumn{1}{l|}{0.64} & \multicolumn{1}{l|}{0.65} & \multicolumn{1}{l|}{0.65} & \multicolumn{1}{l|}{0.65} & \multicolumn{1}{l|}{0.65} & \multicolumn{1}{l|}{0.67} & \multicolumn{1}{l|}{0.67} & \multicolumn{1}{l|}{0.66} & \multicolumn{1}{l|}{0.69} & \multicolumn{1}{l|}{0.68} & 0.67 \\ \hline

\end{tabular}

}
\caption{Performance of different combinations of ML methods and features on every origin  subset.}
\label{tab:result_per_origin}
\end{table}

\begin{itemize}
    \item On the Vent subset, MNB with BoW feature vector outperformed all other models, with an F1-Score of 71\%, followed by MNB with tf-idf and CNN models, with 70\% and 66\% F1-Score, respectively. Here RF was the worst-performing classifier, regardless of the features used.
    \item On ISEAR subset, both MNB+Tfidf and MBN+BoW  achieved highest F1-Scores of 75\% and 74\% respectively, but MNB+BoW however achieving a better recall. Our CNN model achieved an F1-Score of 71\%, and the RF classifier exhibited the worse results.
    \item For the CEASE subset, the MNB classifier trained with Tfidf features achieved an F1-Score of 80\%, followed by GB with tf-idf and the CNN model with 75\% and 74\% F1-Score respectively.
\end{itemize} 

From Table \ref{tab:result_per_origin} we can conclude the following:
\begin{enumerate}
    \item Overall, classifiers with the best hyperparameters tuned on VIC achieved better when trained and tested on the smaller subset (CEASE), and on average, they performed a little better on each subset than in the overall dataset, telling us that the models are fairly good at generalizing and that they are more suitable for short texts classifications. Note that the average sample length in Vent is higher than in ISEAR which is higher than in CEASE.
    \item Every model performed better with Tfidf, compared to BoW feature vectors across all subsets, except for RF ( with virtually the same result regardless of the feature type) and  LR and GB where in general, training with BoW feature vectors.
    \item At first glance, similarly to the classical ML classifiers, for CNN models the shorter the sample texts length the better the performance, while BiLSTM seems unaffected by the sample text length.
\end{enumerate}

\subsection*{Leave-one-out Training and Testing}

When leaving one subset for testing and using the other two for training, as depicted in Table \ref{tab:leaveOneTest}, the results show that models trained on VENT+ISEAR perform better, in almost all instances of train and testing, except the SVM model with Bag-of-Words, which showed a poor F1-score (45\%). These results can naively be explained by the fact that VENT+ISEAR models benefited from having a large number of training
samples(\textasciitilde{}4.5K) and greater average sample length than those in the testing set (CEASE subset).

\begin{table}[!ht]
\centering
\resizebox{0.82\columnwidth}{!}{%
\begin{tabular}{cl|llllllllllll|}
\cline{3-14}
\multicolumn{1}{l}{} & \multicolumn{1}{c|}{\textbf{}} & \multicolumn{12}{c|}{\textbf{Test Set}} \\ \cline{3-14} 
\multicolumn{1}{l}{} &  & \multicolumn{4}{c|}{\textbf{Vent}} & \multicolumn{4}{c|}{\textbf{ISEAR}} & \multicolumn{4}{c|}{\textbf{CEASE}} \\ \hline
\multicolumn{1}{|l|}{\textbf{Feature}} & \textbf{Model} & \multicolumn{1}{l|}{\textbf{Acc}} & \multicolumn{1}{l|}{\textbf{Precision}} & \multicolumn{1}{l|}{\textbf{Recall}} & \multicolumn{1}{l|}{\textbf{F1-Score}} & \multicolumn{1}{l|}{\textbf{Acc}} & \multicolumn{1}{l|}{\textbf{Precision}} & \multicolumn{1}{l|}{\textbf{Recall}} & \multicolumn{1}{l|}{\textbf{F1-Score}} & \multicolumn{1}{l|}{\textbf{Acc}} & \multicolumn{1}{l|}{\textbf{Precision}} & \multicolumn{1}{l|}{\textbf{Recall}} & \textbf{F1-Score} \\ \hline
\multicolumn{1}{|l|}{\multirow{6}{*}{\textbf{tfidf}}} & \textbf{DT} & \multicolumn{1}{l|}{0.56} & \multicolumn{1}{l|}{0.57} & \multicolumn{1}{l|}{0.45} & 0.50 & \multicolumn{1}{l|}{0.57} & \multicolumn{1}{l|}{0.60} & \multicolumn{1}{l|}{0.42} & 0.50 & \multicolumn{1}{l|}{0.65} & \multicolumn{1}{l|}{0.63} & \multicolumn{1}{l|}{0.71} & 0.67 \\ \cline{2-14} 
\multicolumn{1}{|l|}{} & \textbf{RF} & \multicolumn{1}{l|}{0.50} & \multicolumn{1}{l|}{0.50} & \multicolumn{1}{l|}{1.00} & \textbf{0.67} & \multicolumn{1}{l|}{0.50} & \multicolumn{1}{l|}{0.50} & \multicolumn{1}{l|}{1.00} & \textbf{0.67} & \multicolumn{1}{l|}{0.48} & \multicolumn{1}{l|}{0.36} & \multicolumn{1}{l|}{0.06} & 0.11 \\ \cline{2-14} 
\multicolumn{1}{|l|}{} & \textbf{MNB} & \multicolumn{1}{l|}{0.57} & \multicolumn{1}{l|}{0.55} & \multicolumn{1}{l|}{0.82} & 0.66 & \multicolumn{1}{l|}{0.57} & \multicolumn{1}{l|}{0.54} & \multicolumn{1}{l|}{0.89} & \textbf{0.67} & \multicolumn{1}{l|}{0.61} & \multicolumn{1}{l|}{0.57} & \multicolumn{1}{l|}{0.90} & 0.70 \\ \cline{2-14} 
\multicolumn{1}{|l|}{} & \textbf{LR} & \multicolumn{1}{l|}{0.58} & \multicolumn{1}{l|}{0.56} & \multicolumn{1}{l|}{0.66} & 0.61 & \multicolumn{1}{l|}{0.61} & \multicolumn{1}{l|}{0.60} & \multicolumn{1}{l|}{0.67} & 0.63 & \multicolumn{1}{l|}{0.71} & \multicolumn{1}{l|}{0.65} & \multicolumn{1}{l|}{0.87} & \textbf{0.75} \\ \cline{2-14} 
\multicolumn{1}{|l|}{} & \textbf{GB} & \multicolumn{1}{l|}{0.57} & \multicolumn{1}{l|}{0.58} & \multicolumn{1}{l|}{0.49} & 0.53 & \multicolumn{1}{l|}{0.57} & \multicolumn{1}{l|}{0.59} & \multicolumn{1}{l|}{0.48} & 0.53 & \multicolumn{1}{l|}{0.63} & \multicolumn{1}{l|}{0.63} & \multicolumn{1}{l|}{0.60} & 0.62 \\ \cline{2-14} 
\multicolumn{1}{|l|}{} & \textbf{SVC} & \multicolumn{1}{l|}{0.57} & \multicolumn{1}{l|}{0.56} & \multicolumn{1}{l|}{0.63} & 0.59 & \multicolumn{1}{l|}{0.61} & \multicolumn{1}{l|}{0.60} & \multicolumn{1}{l|}{0.65} & 0.62 & \multicolumn{1}{l|}{0.68} & \multicolumn{1}{l|}{0.64} & \multicolumn{1}{l|}{0.86} & 0.73 \\ \hline
\hline
\multicolumn{1}{|l|}{\multirow{6}{*}{\textbf{BoW}}} & \textbf{DT} & \multicolumn{1}{l|}{0.56} & \multicolumn{1}{l|}{0.56} & \multicolumn{1}{l|}{0.50} & 0.53 & \multicolumn{1}{l|}{0.57} & \multicolumn{1}{l|}{0.58} & \multicolumn{1}{l|}{0.48} & 0.53 & \multicolumn{1}{l|}{0.6} & \multicolumn{1}{l|}{0.59} & \multicolumn{1}{l|}{0.62} & 0.60 \\ \cline{2-14} 
\multicolumn{1}{|l|}{} & \textbf{RF} & \multicolumn{1}{l|}{0.50} & \multicolumn{1}{l|}{0.50} & \multicolumn{1}{l|}{1.00} & \textbf{0.67} & \multicolumn{1}{l|}{0.50} & \multicolumn{1}{l|}{0.50} & \multicolumn{1}{l|}{1.00} & \textbf{0.67} & \multicolumn{1}{l|}{0.51} & \multicolumn{1}{l|}{0.57} & \multicolumn{1}{l|}{0.06} & 0.11 \\ \cline{2-14} 
\multicolumn{1}{|l|}{} & \textbf{MNB} & \multicolumn{1}{l|}{0.53} & \multicolumn{1}{l|}{0.52} & \multicolumn{1}{l|}{0.89} & 0.65 & \multicolumn{1}{l|}{0.54} & \multicolumn{1}{l|}{0.52} & \multicolumn{1}{l|}{0.93} & \textbf{0.67} & \multicolumn{1}{l|}{0.52} & \multicolumn{1}{l|}{0.51} & \multicolumn{1}{l|}{0.89} & 0.65 \\ \cline{2-14} 
\multicolumn{1}{|l|}{} & \textbf{LR} & \multicolumn{1}{l|}{0.55} & \multicolumn{1}{l|}{0.56} & \multicolumn{1}{l|}{0.52} & 0.54 & \multicolumn{1}{l|}{0.58} & \multicolumn{1}{l|}{0.60} & \multicolumn{1}{l|}{0.51} & 0.55 & \multicolumn{1}{l|}{0.67} & \multicolumn{1}{l|}{0.63} & \multicolumn{1}{l|}{0.79} & \textbf{0.70} \\ \cline{2-14} 
\multicolumn{1}{|l|}{} & \textbf{GB} & \multicolumn{1}{l|}{0.57} & \multicolumn{1}{l|}{0.64} & \multicolumn{1}{l|}{0.34} & 0.44 & \multicolumn{1}{l|}{0.59} & \multicolumn{1}{l|}{0.67} & \multicolumn{1}{l|}{0.34} & 0.45 & \multicolumn{1}{l|}{0.7} & \multicolumn{1}{l|}{0.75} & \multicolumn{1}{l|}{0.60} & 0.67 \\ \cline{2-14} 
\multicolumn{1}{|l|}{} & \textbf{SVC} & \multicolumn{1}{l|}{0.58} & \multicolumn{1}{l|}{0.61} & \multicolumn{1}{l|}{0.43} & 0.51 & \multicolumn{1}{l|}{0.57} & \multicolumn{1}{l|}{0.61} & \multicolumn{1}{l|}{0.40} & 0.49 & \multicolumn{1}{l|}{0.67} & \multicolumn{1}{l|}{0.67} & \multicolumn{1}{l|}{0.70} & 0.68 \\ \hline

\end{tabular}
}
\caption{Summarization of the best performing classical ML models using different combinations of two subsets for training and one for testing.}
\label{tab:leaveOneTest}
\end{table}

\section*{Error Analysis }
Error analysis is a crucial step in any Natural Language Processing (NLP) study, as it provides insight into the strengths and weaknesses of the model and helps identify areas for improvement. In this section, we present an error analysis of our guilt detection models, in which we analyze the most common types of errors made by our models and attempt to understand the causes of these errors. By examining these errors, we hope to gain a deeper understanding of the nature of the task of detecting guilt in text and to identify potential avenues for improving the performance of our models in future work. A sample of misclassified examples is presented in Table \ref{tab:vic_sample_misclassified}, and each instance of misclassification is explored in detail in this section.

\begin{table}[!ht]
\centering
\resizebox{0.87\columnwidth}{!}{%
 \begin{tabular}{p{0.80\linewidth} | p{0.10\linewidth}| p{0.10\linewidth}}
\multicolumn{1}{c|}{\textbf{Text}} & \multicolumn{1}{c|}{\textbf{Label}} & \multicolumn{1}{c}{\textbf{Predicted}} \\ \hline
I really can’t anymore & no guilt & guilt \\ \hline
me: hey this person hasn't talked to me in a bit my brain: they're dead me: well no they could be driving or with friends or... my brain: they hate you and they're dead me: no -my brain: THEYRE DEAD THEYRE DEAD THEYRE DEAD & no guilt & guilt \\ \hline
My mother had for some time been trying to separate me from a ágood friend who, she thought, was not good company for me. áFinally, at breakfast one day, we had an argument and I tried to ádefend my friend. & no guilt & guilt \\ \hline
A twitter Stan asked me if I have some pills to gargleTried clicking on the notifications and it said they got deleted;) & no guilt & guilt \\ \hline
Disordered eating impliedThis is harder than i anticipated considering how hungry we were ashfkdjsj  camera-emoji & guilt & no guilt \\ \hline
Someone told a lie that I had stolen his money. & no guilt & guilt \\ \hline
Heard that my girl-friend was chosen for the English lectures and áI was not. I lost my temper and she is very upset now. & guilt & no guilt \\ \hline
NSFWSo I should have prepped my ass a lot more than that lmao & guilt & no guilt \\ \hline
When two drug addicts tried to take away my money. & no guilt & guilt \\ \hline
When one of my lovers told me that I was a flirt. & guilt & no guilt \\ \hline
\end{tabular}
}
\caption{ A sample of misclassified examples from dataset VIC, using MNB+Tfidf}
\label{tab:vic_sample_misclassified}
\end{table}

\begin{enumerate}

\item The text "I really can't anymore" seems to express a feeling of exhaustion or despair, which could be interpreted as a negative emotional state and lead to a prediction of guilt. However, without more context, it is difficult to say for certain whether guilt is an appropriate label. It is possible that the model may have been influenced by the word "can't" which could be interpreted as a form of guilt for not being able to handle a situation as could also be possible that the word "anymore" triggered the model to predict guilt, as it suggests some kind of frustration or negative emotion. It is also possible that the model may not have encountered enough examples of this type of language and therefore is not able to accurately predict the label.

\item "me: hey this person hasn't talked to me in a bit my brain: they're dead me: well no they could be driving or with friends or... my brain: they hate you and they're dead me: no -my brain: THEYRE DEAD THEYRE DEAD THEYRE DEAD". The predicted label is guilt while the actual label is no guilt
\begin{enumerate}
    \item The model might have misinterpreted the use of all caps as an expression of strong negative emotions, leading to the prediction of guilt. However, in this context, the person is simply describing an internal dialogue, and there is no actual event that would cause guilt.
    \item The sample seems to be a humorous or exaggerated representation of intrusive or anxious thoughts, so the predicted label of guilt may indicate that the speaker feels guilty for having these thoughts or for being anxious about their relationship with the person who has not contacted them.
\end{enumerate}

\item "My mother had for some time been trying to separate me from a good friend who, she thought, was not good company for me. Finally, at breakfast one day, we had an argument, and I tried to defend my friend." - The predicted label is guilt, while the actual label is no guilt. 
\begin{enumerate}
    \item The text describes a disagreement between the speaker and their mother over whether their friend is a good influence, but there is no indication of the speaker feeling guilty.
    \item It is possible that the model picked up on the word "argument" and predicted guilt based on the assumption that arguments are usually associated with negative emotions. However, in this context, the person is simply defending their friend, and there is no indication of guilt.
    \item This sample describes a conflict between the speaker and their mother over a friend. The predicted label of guilt may indicate that the speaker feels guilty for not being able to resolve the conflict or for causing their mother stress.
\end{enumerate}

\item "A twitter Stan asked me if I have some pills to gargle...Tried clicking on the notifications and it said they got deleted;)": The true label is "no guilt" and the predicted label is "guilt". This sample is difficult to interpret without more context, but it may indicate that the speaker is being asked for a favor or being made an uncomfortable request. The predicted label of guilt may indicate that the speaker feels guilty for not being able to fulfill the request, but it is unclear why the model predicted guilt in this instance, as there is no indication of negative emotions or events.

\item "Disordered eating implied This is harder than i anticipated considering how hungry we were ashfkdjsj camera-emoji": The true label is "no guilt" and the predicted label is "no guilt". 
\begin{enumerate}
    \item The text expresses frustration about hunger, but there is no indication of guilt.
    \item This sample seems to indicate that the speaker is struggling with disordered eating and finding it difficult to manage their hunger. The predicted label of no guilt may indicate that the speaker does not feel guilty about their struggle with disordered eating or their difficulty managing their hunger.
    \item It is possible that the model did not pick up on the implication of disordered eating in this text, leading to misclassification.
\end{enumerate}

\item "Someone told a lie that I had stolen his money." - The predicted label is guilt while the actual label is no guilt. It is possible that the model picked up on the word "stolen" and predicted guilt based on the assumption that stealing is associated with negative emotions. However, in this context, the person is simply reporting an accusation.

\item "Heard that my girl-friend was chosen for the English lectures and I was not. I lost my temper and she is very upset now.": The predicted label is "no guilt", but the true label is "guilt". 
\begin{enumerate}
    \item It's possible that the model did not pick up on the guilt in the text, which is implied through the author's loss of temper and the upset caused to the girlfriend.
    \item  The model may have also been confused by the use of the word "good" in the text, which could have made it think that the text was more related to "no guilt" than "guilt."
    \item The model may have failed to pick up on the implication that the person's loss of temper was a negative event, leading to misclassification.
\end{enumerate}

\item "NSFWSo I should have prepped my ass a lot more than that lmao" - The predicted label is no guilt while the actual label is guilt. 

\begin{enumerate}
    \item Here, the model could not pick up on the implication of a negative or embarrassing event, leading to misclassification.
    \item  A second possibility is that the model did not pick up on the guilt in the text, which is implied through the author's use of a phrase that suggests a lack of preparation. 
    \item The model may have also been confused by the use of the NSFW tag (usually used to implicate something of mature or sexually explicit nature), which could have made it think that the text was more related to "no guilt" than "guilt".
\end{enumerate}

\item "When two drug addicts tried to take away my money." - The predicted label is guilt while the actual label is no guilt. 
\begin{enumerate}
    \item  The model may have incorrectly predicted guilt for this text due to the mention of "drug addicts" and "taking away my money." It's possible that the model assumed the author must have done something wrong to have encountered this situation when they were simply the victim of an attempted robbery.
    \item Ambiguity: The text is brief and lacks a lot of context, so it's possible that the author did something that could be interpreted as guilt-inducing in this situation. Perhaps they were carrying a large amount of money and were in an area known for drug activity, or maybe they responded to the situation with aggression or violence. And the lack of further context may have led the model to fail.
\end{enumerate}

\item "When one of my lovers told me that I was a flirt" - The predicted label is ni guilt while the actual label is  guilt.  Possible reasons for the misclassification could be:
\begin{enumerate}
    \item Ambiguity in the text: The phrase "one of my lovers" could indicate that the person has multiple partners, which might be interpreted as flirting. The classifier might have assigned a guilt label based on this ambiguity, while in reality, the person might not have been flirting at all.
    \item Lack of context: The classifier might not have had enough information about the situation to judge accurately. The text doesn't provide details about the nature of the relationship or the context of the accusation, which could have influenced the label.
    \item Personal bias: The classifier might be influenced by personal bias or assumptions about what constitutes "flirting". Due to the training data, the model might have learned a biased interpretation of the term "flirt" that doesn't align with the writer's intentions.
\end{enumerate}

In general, misclassification can occur for various reasons, including ambiguity in the text, errors in the training data or model, and differences in how people interpret or understand language. It's important to continually evaluate and improve models to minimize misclassifications and improve accuracy. \\

\end{enumerate}

\section*{Conclusion}
In this study, we developed a new binary classification dataset for guilt detection by extracting guilt and non-guilt samples from three existing multi-class emotion detection datasets. By doing so, we provided a unique resource for researchers and practitioners to study and develop machine learning models for guilt detection, a critical application in the fields of studies at the intersection of psychology and computational linguistics. We trained classical machine learning models and two neural network models, CNN and BiLSTM, to establish baselines using both Tfidf and BoW feature vectors.

Our results showed that, in general, every model performed better with Tfidf compared to BoW feature vectors, except for RF. LR and GB achieved better results when trained with BoW feature vectors. Moreover, CNN models performed better with shorter sample text lengths, while BiLSTM models were not affected by the text length.

Regarding subset-specific performance, on the Vent subset, the MNB model with BoW features achieved the highest F1-Score of 71\%, outperforming all other models. On the ISEAR subset, both MNB with Tfidf and MNB with BoW achieved the highest F1-Scores of 75\% and 74\%, respectively, with the latter achieving a better recall. On the CEASE subset, the MNB classifier trained with Tfidf features achieved the highest F1-Score of 80\%, followed by GB and the CNN model, achieving F1-Scores of 75%.

Overall, our study provides a valuable benchmark for future research in guilt detection. Our error analysis also revealed some of the weaknesses of our models, which could be addressed in future studies to improve performance. Our work highlights the importance of carefully selecting the training data and the choice of machine learning models when developing systems for guilt detection. Furthermore, our leave-one-out analysis demonstrated that our models can generalize to new data with reasonable accuracy.

\bibliography{main}

\begin{thebibliography}{10}
\urlstyle{rm}
\expandafter\ifx\csname url\endcsname\relax
  \def\url#1{\texttt{#1}}\fi
\expandafter\ifx\csname urlprefix\endcsname\relax\def\urlprefix{URL }\fi
\expandafter\ifx\csname doiprefix\endcsname\relax\def\doiprefix{DOI: }\fi
\providecommand{\bibinfo}[2]{#2}
\providecommand{\eprint}[2][]{\url{#2}}

\bibitem{rime1991}
\bibinfo{author}{Rimé, B.}, \bibinfo{author}{Mesquita, B.},
  \bibinfo{author}{Boca, S.} \& \bibinfo{author}{Philippot, P.}
\newblock \bibinfo{journal}{\bibinfo{title}{Beyond the emotional event: Six
  studies on the social sharing of emotion}}.
\newblock {\emph{\JournalTitle{Cognition and Emotion}}}
  \textbf{\bibinfo{volume}{5}}, \bibinfo{pages}{435--465},
  \doiprefix\url{10.1080/02699939108411052} (\bibinfo{year}{1991}).
\newblock \eprint{https://doi.org/10.1080/02699939108411052}.

\bibitem{bergermilkman2009}
\bibinfo{author}{Berger, J.} \& \bibinfo{author}{Milkman, K.}
\newblock \bibinfo{journal}{\bibinfo{title}{What makes online content viral}}.
\newblock {\emph{\JournalTitle{Journal of Marketing Research}}}
  \textbf{\bibinfo{volume}{49}}, \bibinfo{pages}{192--205},
  \doiprefix\url{10.2139/ssrn.1528077} (\bibinfo{year}{2009}).

\bibitem{rawlings1970a}
\bibinfo{author}{Rawlings, E.~I.}
\newblock \bibinfo{title}{Reactive guilt and anticipatory guilt in altruistic
  behavior}.
\newblock In \bibinfo{editor}{Macaulay, J.} \& \bibinfo{editor}{Berkowitz, L.}
  (eds.) \emph{\bibinfo{booktitle}{Altruism and Helping Behavior}},
  \bibinfo{pages}{163–177} (\bibinfo{publisher}{Academic Press},
  \bibinfo{address}{New York}, \bibinfo{year}{1970}).

\bibitem{Kugler1992OnCA}
\bibinfo{author}{Kugler, K.~E.} \& \bibinfo{author}{Jones, W.~H.}
\newblock \bibinfo{journal}{\bibinfo{title}{On conceptualizing and assessing
  guilt.}}
\newblock {\emph{\JournalTitle{Journal of Personality and Social Psychology}}}
  \textbf{\bibinfo{volume}{62}}, \bibinfo{pages}{318--327}
  (\bibinfo{year}{1992}).

\bibitem{treeby2021}
\bibinfo{author}{Kealy, D.}, \bibinfo{author}{Treeby, M.~S.} \&
  \bibinfo{author}{Rice, S.~M.}
\newblock \bibinfo{journal}{\bibinfo{title}{Shame, guilt, and suicidal
  thoughts: The interaction matters}}.
\newblock {\emph{\JournalTitle{British Journal of Clinical Psychology}}}
  \textbf{\bibinfo{volume}{60}}, \bibinfo{pages}{414--423},
  \doiprefix\url{https://doi.org/10.1111/bjc.12291} (\bibinfo{year}{2021}).
\newblock
  \eprint{https://bpspsychub.onlinelibrary.wiley.com/doi/pdf/10.1111/bjc.12291}.

\bibitem{tangney2002a}
\bibinfo{author}{Tangney, J.} \& \bibinfo{author}{Dearing, R.}
\newblock \emph{\bibinfo{title}{Shame and guilt}} (\bibinfo{publisher}{Guilford
  Press}, \bibinfo{address}{New York}, \bibinfo{year}{2002}).

\bibitem{bryan2015}
\bibinfo{author}{Bryan, C.~J.} \emph{et~al.}
\newblock \bibinfo{journal}{\bibinfo{title}{Guilt as a mediator of the
  relationship between depression and posttraumatic stress with suicide
  ideation in two samples of military personnel and veterans}}.
\newblock {\emph{\JournalTitle{International Journal of Cognitive Therapy}}}
  \textbf{\bibinfo{volume}{8}}, \bibinfo{pages}{143--155},
  \doiprefix\url{10.1521/ijct.2015.8.2.143} (\bibinfo{year}{2015}).
\newblock \eprint{https://doi.org/10.1521/ijct.2015.8.2.143}.

\bibitem{CUNNINGHAM2017227}
\bibinfo{author}{Cunningham, K.~C.} \emph{et~al.}
\newblock \bibinfo{journal}{\bibinfo{title}{A model comparison approach to
  trauma-related guilt as a mediator of the relationship between ptsd symptoms
  and suicidal ideation among veterans}}.
\newblock {\emph{\JournalTitle{Journal of Affective Disorders}}}
  \textbf{\bibinfo{volume}{221}}, \bibinfo{pages}{227--231},
  \doiprefix\url{https://doi.org/10.1016/j.jad.2017.06.046}
  (\bibinfo{year}{2017}).

\bibitem{balahur-etal-2011-detecting}
\bibinfo{author}{Balahur, A.}, \bibinfo{author}{Hermida, J.~M.} \&
  \bibinfo{author}{Montoyo, A.}
\newblock \bibinfo{title}{Detecting implicit expressions of sentiment in text
  based on commonsense knowledge}.
\newblock In \emph{\bibinfo{booktitle}{Proceedings of the 2nd Workshop on
  Computational Approaches to Subjectivity and Sentiment Analysis ({WASSA}
  2.011)}}, \bibinfo{pages}{53--60} (\bibinfo{publisher}{Association for
  Computational Linguistics}, \bibinfo{address}{Portland, Oregon},
  \bibinfo{year}{2011}).

\bibitem{ghosh-etal-2020-cease}
\bibinfo{author}{Ghosh, S.}, \bibinfo{author}{Ekbal, A.} \&
  \bibinfo{author}{Bhattacharyya, P.}
\newblock \bibinfo{title}{{{CEASE}, a Corpus of Emotion Annotated Suicide notes
  in {E}nglish}}.
\newblock In \emph{\bibinfo{booktitle}{Proceedings of the Twelfth Language
  Resources and Evaluation Conference}}, \bibinfo{pages}{1618--1626}
  (\bibinfo{publisher}{European Language Resources Association},
  \bibinfo{address}{Marseille, France}, \bibinfo{year}{2020}).

\bibitem{balahur-etal-2012-expanding}
\bibinfo{author}{Balahur, A.}, \bibinfo{author}{Hermida, J.~M.} \&
  \bibinfo{author}{Montoyo, A.}
\newblock \bibinfo{journal}{\bibinfo{title}{Building and exploiting emotinet, a
  knowledge base for emotion detection based on the appraisal theory model}}.
\newblock {\emph{\JournalTitle{IEEE Transactions on Affective Computing}}}
  \textbf{\bibinfo{volume}{3}}, \bibinfo{pages}{88--101},
  \doiprefix\url{10.1109/T-AFFC.2011.33} (\bibinfo{year}{2012}).

\bibitem{Lykousas_2019}
\bibinfo{author}{Lykousas, N.}, \bibinfo{author}{Patsakis, C.},
  \bibinfo{author}{Kaltenbrunner, A.} \& \bibinfo{author}{G{\'{o}}mez, V.}
\newblock \bibinfo{journal}{\bibinfo{title}{{Sharing Emotions at Scale: The
  Vent Dataset}}}.
\newblock {\emph{\JournalTitle{Proceedings of the International {AAAI}
  Conference on Web and Social Media}}} \textbf{\bibinfo{volume}{13}},
  \bibinfo{pages}{611--619}, \doiprefix\url{10.1609/icwsm.v13i01.3361}
  (\bibinfo{year}{2019}).

\bibitem{ohman-etal-2020-xed}
\bibinfo{author}{{\"O}hman, E.}, \bibinfo{author}{P{\`a}mies, M.},
  \bibinfo{author}{Kajava, K.} \& \bibinfo{author}{Tiedemann, J.}
\newblock \bibinfo{title}{{XED}: A multilingual dataset for sentiment analysis
  and emotion detection}.
\newblock In \emph{\bibinfo{booktitle}{Proceedings of the 28th International
  Conference on Computational Linguistics}}, \bibinfo{pages}{6542--6552},
  \doiprefix\url{10.18653/v1/2020.coling-main.575}
  (\bibinfo{publisher}{International Committee on Computational Linguistics},
  \bibinfo{address}{Barcelona, Spain (Online)}, \bibinfo{year}{2020}).

\bibitem{demszky-etal-2020-goemotions}
\bibinfo{author}{Demszky, D.} \emph{et~al.}
\newblock \bibinfo{title}{{G}o{E}motions: A dataset of fine-grained emotions}.
\newblock In \emph{\bibinfo{booktitle}{Proceedings of the 58th Annual Meeting
  of the Association for Computational Linguistics}},
  \bibinfo{pages}{4040--4054}, \doiprefix\url{10.18653/v1/2020.acl-main.372}
  (\bibinfo{publisher}{Association for Computational Linguistics},
  \bibinfo{address}{Online}, \bibinfo{year}{2020}).

\bibitem{hsu-etal-2018-emotionlines}
\bibinfo{author}{Hsu, C.-C.}, \bibinfo{author}{Chen, S.-Y.},
  \bibinfo{author}{Kuo, C.-C.}, \bibinfo{author}{Huang, T.-H.} \&
  \bibinfo{author}{Ku, L.-W.}
\newblock \bibinfo{title}{{E}motion{L}ines: An emotion corpus of multi-party
  conversations}.
\newblock In \emph{\bibinfo{booktitle}{Proceedings of the Eleventh
  International Conference on Language Resources and Evaluation ({LREC} 2018)}}
  (\bibinfo{publisher}{European Language Resources Association (ELRA)},
  \bibinfo{address}{Miyazaki, Japan}, \bibinfo{year}{2018}).

\bibitem{Scherer_1994}
\bibinfo{author}{Scherer, K.~R.} \& \bibinfo{author}{Wallbott, H.~G.}
\newblock \bibinfo{journal}{\bibinfo{title}{{Evidence for universality and
  cultural variation of differential emotion response patterning:
  Correction.}}}
\newblock {\emph{\JournalTitle{Journal of Personality and Social Psychology}}}
  \textbf{\bibinfo{volume}{67}}, \bibinfo{pages}{55--55},
  \doiprefix\url{10.1037/0022-3514.67.1.55} (\bibinfo{year}{1994}).

\bibitem{troiano2019crowdsourcing}
\bibinfo{author}{Troiano, E.}, \bibinfo{author}{Pad{\'o}, S.} \&
  \bibinfo{author}{Klinger, R.}
\newblock \bibinfo{journal}{\bibinfo{title}{Crowdsourcing and validating
  event-focused emotion corpora for german and english}}.
\newblock {\emph{\JournalTitle{arXiv preprint arXiv:1905.13618}}}
  (\bibinfo{year}{2019}).

\bibitem{ekman1971a}
\bibinfo{author}{Ekman, P.}
\newblock \bibinfo{title}{Universals and cultural differences in facial
  expressions of emotion}.
\newblock In \emph{\bibinfo{booktitle}{Nebraska symposium on motivation}}
  (\bibinfo{publisher}{University of Nebraska Press}, \bibinfo{year}{1971}).

\bibitem{Ekman:1992a}
\bibinfo{author}{Ekman, P.}
\newblock \bibinfo{journal}{\bibinfo{title}{An argument for basic emotions}}.
\newblock {\emph{\JournalTitle{Cognition and Emotion}}}
  \textbf{\bibinfo{volume}{6}}, \bibinfo{pages}{169--200},
  \doiprefix\url{10.1080/02699939208411068} (\bibinfo{year}{1992}).
\newblock \eprint{https://doi.org/10.1080/02699939208411068}.

\bibitem{plutchik1980a}
\bibinfo{author}{Plutchik, R.}
\newblock \bibinfo{journal}{\bibinfo{title}{A general psychoevolutionary theory
  of emotion}}.
\newblock {\emph{\JournalTitle{Theories of emotion}}}
  \textbf{\bibinfo{volume}{1}}, \bibinfo{pages}{3–31} (\bibinfo{year}{1980}).

\bibitem{Nurudin_2021}
\bibinfo{author}{Alvarez{-}Gonzalez, N.}, \bibinfo{author}{Kaltenbrunner, A.}
  \& \bibinfo{author}{G{\'{o}}mez, V.}
\newblock \bibinfo{journal}{\bibinfo{title}{Uncovering the limits of text-based
  emotion detection}}.
\newblock {\emph{\JournalTitle{CoRR}}}
  \textbf{\bibinfo{volume}{abs/2109.01900}} (\bibinfo{year}{2021}).
\newblock \eprint{2109.01900}.

\bibitem{kim-2014-convolutional}
\bibinfo{author}{Kim, Y.}
\newblock \bibinfo{title}{Convolutional neural networks for sentence
  classification}.
\newblock In \emph{\bibinfo{booktitle}{Proceedings of the 2014 Conference on
  Empirical Methods in Natural Language Processing ({EMNLP})}},
  \bibinfo{pages}{1746--1751}, \doiprefix\url{10.3115/v1/D14-1181}
  (\bibinfo{publisher}{Association for Computational Linguistics},
  \bibinfo{address}{Doha, Qatar}, \bibinfo{year}{2014}).

\bibitem{hochreiter1997a}
\bibinfo{author}{Hochreiter, S.} \& \bibinfo{author}{Schmidhuber, J.}
\newblock \bibinfo{journal}{\bibinfo{title}{Long shortterm memory}}.
\newblock {\emph{\JournalTitle{Neural computation}}}
  \textbf{\bibinfo{volume}{9}}, \bibinfo{pages}{1735–1780}.

\bibitem{cho-etal-2014-properties}
\bibinfo{author}{Cho, K.}, \bibinfo{author}{van Merri{\"e}nboer, B.},
  \bibinfo{author}{Bahdanau, D.} \& \bibinfo{author}{Bengio, Y.}
\newblock \bibinfo{title}{On the properties of neural machine translation:
  Encoder{--}decoder approaches}.
\newblock In \emph{\bibinfo{booktitle}{Proceedings of {SSST}-8, Eighth Workshop
  on Syntax, Semantics and Structure in Statistical Translation}},
  \bibinfo{pages}{103--111}, \doiprefix\url{10.3115/v1/W14-4012}
  (\bibinfo{publisher}{Association for Computational Linguistics},
  \bibinfo{address}{Doha, Qatar}, \bibinfo{year}{2014}).

\bibitem{fazl_etal_hope_2022}
\bibinfo{author}{Balouchzahi, F.}, \bibinfo{author}{Sidorov, G.} \&
  \bibinfo{author}{Gelbukh, A.}
\newblock \bibinfo{title}{Polyhope: Two-level hope speech detection from
  tweets}, \doiprefix\url{10.48550/ARXIV.2210.14136} (\bibinfo{year}{2022}).

\bibitem{fazl_etall_guilt_2022}
\bibinfo{author}{Balouchzahi, F.}, \bibinfo{author}{Butt, S.},
  \bibinfo{author}{Sidorov, G.} \& \bibinfo{author}{Gelbukh, A.}
\newblock \bibinfo{title}{Reddit: Regret detection and domain identification
  from text}, \doiprefix\url{10.48550/ARXIV.2212.07549} (\bibinfo{year}{2022}).

\bibitem{Schuster_1997}
\bibinfo{author}{Schuster, M.} \& \bibinfo{author}{Paliwal, K.}
\newblock \bibinfo{journal}{\bibinfo{title}{{Bidirectional recurrent neural
  networks}}}.
\newblock {\emph{\JournalTitle{{IEEE} Transactions on Signal Processing}}}
  \textbf{\bibinfo{volume}{45}}, \bibinfo{pages}{2673--2681},
  \doiprefix\url{10.1109/78.650093} (\bibinfo{year}{1997}).

\end{thebibliography}

\section*{Acknowledgements}

The work was done with partial support from the Mexican Government through the grant A1-S-47854 of CONACYT,
Mexico, grants 20220852 and 20220859 of the Secretaría de Investigación y Posgrado of the Instituto Politécnico
Nacional, Mexico. The authors thank the CONACYT for the computing resources brought to them through the
Plataforma de Aprendizaje Profundo para Tecnologías del Lenguaje of the Laboratorio de Supercómputo of the INAOE,
Mexico and acknowledge the support of Microsoft through the Microsoft Latin America Ph.D. Award.

\section*{Author contributions statement}

A.M. conceived and conducted the experiment(s), analyzed the results, and conducted error analysis; N.H. conducted the literature review and analyzed the results; and A.G. and G.S. supervised the entire research.  All authors reviewed the manuscript. 

\section*{Data availability}
The ISEAR dataset is publicly available at https://www.unige.ch/cisa/index.php/download\_file/view/395/296/ and the source paper introducing this dataset is cited correctly in this manuscript. The CEASE dataset is available upon reasonable request to the authors of the paper that first presented it, and a form for requesting access is available at https://www.iitp.ac.in/~ai-nlp-ml/resources.html. Portions of the Vent Dataset are publicly available, while the texts are only available upon reasonable request to the paper's corresponding author who first introduced it. The resulting dataset used in this study, which only contains readers and labels, is also available upon reasonable request due to compliance constraints on some of the samples because of their origin.
 
\section{Ethics statement} 
The research presented in this manuscript involves using three different datasets, namely the ISEAR, CEASE, and Vent datasets. All ethical considerations for each dataset were addressed in each paper that initially introduced them. The dataset comprises texts of three types, social media posts, suicide notes, and psychological study questionnaire answers. The resulting dataset used in this study only contains texts and labels, with no identifying or sensitive information. The authors obtained access to the datasets through reasonable requests to each of the dataset's authors/creators and fulfilled and followed all the compliance requests set forth by the authors. The authors also ensured that identifying and sensitive information was removed from the resulting dataset to protect the individual's privacy. The resulting dataset is available upon reasonable request due to compliance constraints on some of the samples because of their origin.

\section{Competing interests}

The authors declare no competing interests.

\end{document}